\documentclass[letterpaper]{article} 
\usepackage{aaai25} 
\usepackage{times}  
\usepackage{helvet}  
\usepackage{courier}  
\usepackage[hyphens]{url}  
\usepackage{graphicx} 
\usepackage{subfigure}
\usepackage{natbib}  
\usepackage{caption} 
\usepackage{algorithm}
\usepackage{algorithmic}
\usepackage[table]{xcolor}
\usepackage{multirow}
\usepackage[utf8]{inputenc} 
\usepackage[T1]{fontenc}    
\usepackage{booktabs}       
\usepackage{amsfonts}       
\usepackage{microtype}      
\usepackage{xcolor}         
\usepackage{bbding}
\usepackage{amsmath}
\usepackage{amssymb}
\usepackage{mathtools}
\usepackage{amsthm}
\usepackage{newfloat}
\usepackage{listings}
\DeclareCaptionStyle{ruled}{labelfont=normalfont,labelsep=colon,strut=off} 
\floatstyle{ruled}
\newfloat{listing}{tb}{lst}{}
\floatname{listing}{Listing}
\title{Mitigating Prior Shape Bias in Point Clouds via Differentiable Center Learning}
\author{
   \textsuperscript{\rm 1}Zhe Li, , \textsuperscript{\rm 1}Xiying Wang, \textsuperscript{\rm 1}Jinglin Zhao, \textsuperscript{\rm 1}Zheng Wang,  \textsuperscript{\rm 2}Debin Liu, 
	\\\textsuperscript{\rm 2}Laurence T. Yang*
}
\affiliations{
    \textsuperscript{\rm 1}Huazhong University of Science and Technology\\

    \textsuperscript{\rm 2}Zhengzhou University
}

\usepackage{bibentry}

\begin{document}

\maketitle
\begin{figure}[ht]
	\centering
	\subfigure[Traditional center point acquisition method.] {\includegraphics[width=.455\textwidth]{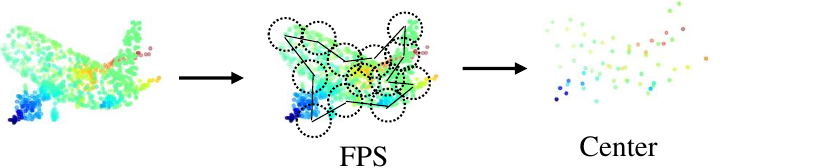}\label{mota}}
	\subfigure[Our proposed center point acquisition method.] {\includegraphics[width=.405\textwidth]{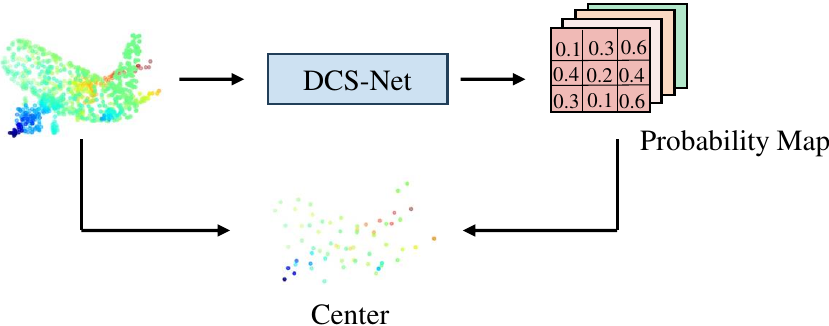}\label{motb}}
	\caption{\textbf{Different center point acquisition methods.} (a) Conventional point cloud models employ the farthest distance sampling method for center point acquisition, resulting in trivial proxy tasks. (b) We propose a differentiable center sampling network (DCS-Net) for center point acquisition and reducing shape prior bias. By computing probability maps and weighting the points accordingly, our approach fully incorporates the semantic information of the point cloud.}
	\label{motivation}
\end{figure}
\begin{abstract}
Masked autoencoding and generative pretraining have achieved remarkable success in computer vision and natural language processing, and more recently, they have been extended to the domain of point cloud modeling. Nevertheless, existing point cloud models suffer from the issue of prior shape bias due to the pre-sampling of center points, which leads to trivial proxy tasks for modeling. These approaches primarily focus on local feature reconstruction, limiting their ability to capture global patterns within point clouds.
In this paper, we argue that the reduced difficulty of proxy tasks hampers the model's capacity to learn expressive representations. To address these limitations, we introduce a novel solution called the Differentiable Center Sampling Network (\textbf{DCS-Net}). It tackles the prior shape bias problem by incorporating both global feature reconstruction and local feature reconstruction as non-trivial proxy tasks, enabling simultaneous learning of both the global and local patterns within point clouds. Experimental results demonstrate that our method enhances the expressive capacity of existing point cloud models and effectively mitigates prior shape bias.
\end{abstract}

\section{Introduction}
\label{introduction}
Learning rich feature representations from unannotated data has emerged as a prominent trend in the field of deep learning, commonly referred to as self-supervised learning. This approach typically involves pretraining models on different proxy tasks and finetuning them on downstream tasks. Self-supervised learning has propelled advancements in natural language processing \cite{brown2020language, kenton2019bert, joshi2020spanbert, liu2019roberta, radford2019language} and computer vision \cite{bao2021beit, chen2020generative, dosovitskiy2020image, touvron2021training, xie2021self, zhu2020deformable} by leveraging the absence of labeled data. The ability to learn meaningful feature representations from data is attributed to the rationality of proxy tasks, such as the masked language modeling and next sentence prediction in BERT \cite{kenton2019bert}, as well as the masked image modeling in MAE \cite{he2022masked}. These proxy tasks provide valuable supervision signals that facilitate the learning of effective representations without explicit labels.

Point clouds play a pivotal role as fundamental data structures in various domains, including autonomous driving and robotics. As a result, point cloud representation learning and generation have gained escalating importance. Existing transformer-based point cloud models \cite{yu2022point, chen2023pointgpt, pang2022masked, liu2022masked, li2023general} commonly employ the farthest point sampling (FPS) \cite{qi2017pointnet++} algorithm to perform center point sampling on the input point cloud, depicted in Figure \ref{mota}. Subsequently, they leverage local feature reconstruction as a proxy task to enable the model to learn the local patterns of the point cloud, which are then finetuned in downstream tasks. However, if we perform center sampling as a separate operation before each training or fine-tuning process, it introduces unnecessary complexity and exhibits two overlooked issues: (i) \textbf{Prior Shape Bias:} the objective of the FPS algorithm is to select a set of points with the maximum average distance as the sampling result, aiming to cover the spatial range of the entire point cloud as much as possible. However, in existing point cloud models, the input is not The original point cloud but rather the center points obtained through FPS sampling. This introduces the prior shape bias as the model has prior knowledge of the approximate shape of the point cloud; (ii) \textbf{Trivial Proxy Task:} center point sampling with the FPS algorithm reduces the difficulty of model pretraining. Since the process of sampling center points is non-differentiable, the model can only use local feature reconstruction as a proxy task during pretraining, focusing solely on learning the local patterns of the point cloud while neglecting global patterns. 

However, proxy tasks for pretraining should be non-trivial. By employing a differentiable approach for center point acquisition, it becomes possible to incorporate both global feature reconstruction and local feature reconstruction as proxy tasks simultaneously.

To address these challenges, we introduce DCS-Net, a differentiable center point acquisition method for point clouds depicted in Figure \ref{motb}. This paradigm shift alters the traditional point cloud model pretraining paradigm (FPS + model pretraining), resulting in a more coherent training process and resolving the issue of prior shape bias. Specifically, we map the point cloud onto a canonical sphere, such that semantically similar parts of objects belonging to the same category are mapped to the same position. On the canonical sphere, we compute a probability map representing the likelihood of each point belonging to a certain category. By weighting all points using this probability map, we obtain the same number of centers as the number of categories. And during the model pretraining phase, we replace the FPS algorithm with DCS-Net for center sampling while keeping the subsequent operations unchanged.

Center points encapsulate rich global information of the point cloud, motivating us to design center point reconstruction as a novel proxy task for pretraining. Existing point cloud models employ unique proxy tasks, such as masked point prediction in Point-BERT and autoregressive generation in PointGPT. However, these tasks primarily focus on local feature reconstruction, overlooking the global patterns within the point cloud. By utilizing DCS-Net to acquire center points, we introduce an additional proxy task, global feature reconstruction, to complement the existing local feature reconstruction in point cloud models. This combined approach enables simultaneous learning of both global and local patterns within the point cloud. With the integration of global feature reconstruction, our method enhances the representation learning of point clouds by capturing both the fine-grained local details and the broader global structures.

To validate the effectiveness of DCS-Net, we conduct extensive experiments on various self-supervised point cloud models and multimodal point cloud models \cite{qi2023contrast, dong2022autoencoders}. During the pretraining process, we replace the FPS algorithm with DCS-Net for center point acquisition and simultaneously perform global feature reconstruction and local feature reconstruction tasks. The experimental results demonstrate the efficacy of our \textbf{DCS-Net+point cloud model} pretraining paradigm. Our approach not only successfully mitigates prior shape bias in the existing point cloud model but also enhances the model capacity in learning expressive representations.
\begin{figure*}
	\centering 
	\includegraphics[scale=0.80]{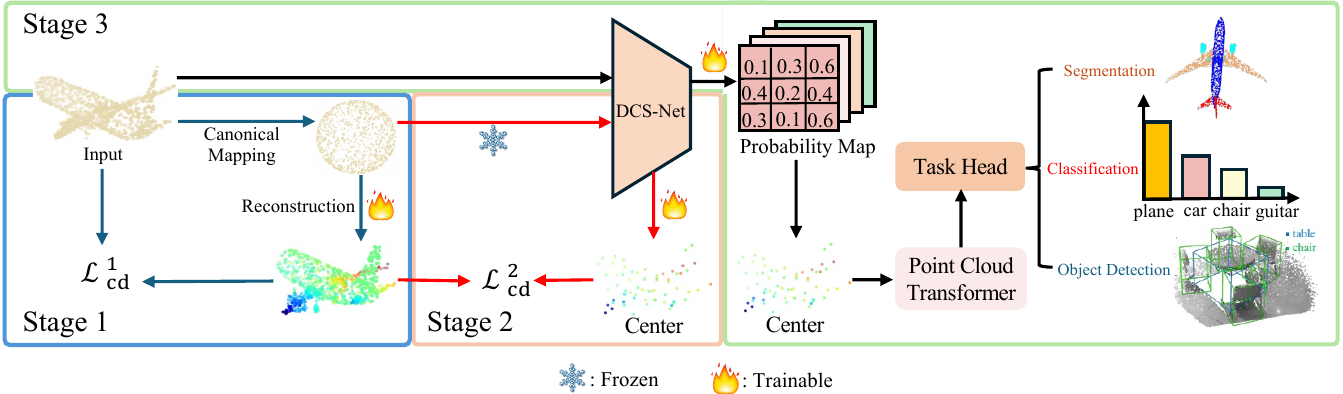} 
	\caption{\textbf{Overview of DCS-Net.} We divide the entire process into three stages. Firstly, we learn a mapping function to a canonical sphere, resulting in a more uniform distribution of semantic information (\textcolor{blue}{represented by the blue arrows}). Next, we employ DCS-Net on the canonical sphere for center point sampling, enabling the initial training of DCS-Net through center point reconstruction (\textcolor{red}{represented by the red arrows}). Finally, we unify the training of DCS-Net and the point cloud model by incorporating global feature reconstruction and local feature reconstruction as proxy tasks. The pretrained model is further finetuned on downstream datasets (represented by the black arrows).}  
	\label{framework} 
\end{figure*}
\section{Related Work}
\label{related}
\subsection{Self-supervised Learning for Point Cloud}
The efficacy of self-supervised learning (SSL) in natural language processing (NLP) and computer vision has inspired researchers to extend SSL frameworks to point cloud representation learning.
Among these methods, contrastive approaches \cite{xie2020pointcontrast, zhang2021self, yang2018foldingnet, navaneet2020image, jing2020self} have received considerable attention. DepthContrast \cite{zhang2021self} generates augmented depth maps and utilizes an instance discrimination task to learn global features. Similarly, MVIF \cite{navaneet2020image} employs cross-modal and cross-view invariance constraints to facilitate self-supervised learning of modal- and view-invariant features. Another direction of research \cite{dong2022autoencoders, qi2023contrast, xue2023ulip, xue2023ulip2} aims to integrate cross-modal information and leverage knowledge transfer from language or image models to enhance 3D learning.
ACT \cite{dong2022autoencoders} introduces cross-modal autoencoders as teacher models to leverage knowledge from other modalities. Recon \cite{qi2023contrast} leverages ensemble distillation to learn from both generative modeling teachers and single/cross-modal contrastive teachers. In the field of generative modeling, \cite{li2018so,achlioptas2018learning,sauder2019self,min2022voxel,yu2022point,zhang2022point} have made significant contributions. Point-BERT \cite{yu2022point} employs masked point prediction as a proxy task, aiming to recover the original point tokens at the masked locations under the supervision of point tokens obtained from the tokenizer. Point-MAE \cite{pang2022masked} extends MAE by randomly masking point patches and reconstructing the masked regions. Point-M2AE \cite{zhang2022point} further incorporates a hierarchical transformer architecture and devises corresponding masking strategies. PointGPT \cite{chen2023pointgpt} designs a transformer-based extractor-generator with a dual masking strategy, aiming to predict the next one in an auto-regressive manner. Different from above, GPM \cite{li2023general} proposes a new pretraining paradigm, seamlessly integrates autoencoding and autoregressive tasks in a point cloud transformer. However, these point cloud models primarily focus on training the model through local feature reconstruction, overlooking the global patterns within the point cloud. Additionally, the non-differentiability of center point acquisition has led to prior shape bias, reducing the training difficulty of the model.
\subsection{Point Cloud Correspondence Learning}
Given a pair of source and target instances, point cloud correspondence learning aims to find corresponding points in the target instance for each point in the source instance. Several existing methods \cite{chen2019edgenet, choy2019fully, gojcic2019perfect} address this task through point cloud registration, utilizing labeled pairwise correspondence as supervision. To alleviate the reliance on explicit supervision, \cite{bhatnagar2020combining} predicts part correspondences to a template using implicit functions, albeit requiring part labels for training. \cite{chen2020unsupervised} proposes a method for unsupervised learning of 3D dense correspondence by leveraging consistent 3D structure points across different instances. However, their model assumes structural similarity among instances, neglecting intra-class variations and the detection of non-existing correspondences between dissimilar shapes within the same category.

In the field of unsupervised learning, \cite{liu2020learning} introduces an unsupervised approach that utilizes part features learned by BAE-NET \cite{chen2019bae} to establish dense correspondences. This method can estimate a confidence score representing the likelihood of correspondence. In contrast, \cite{cheng2022autoregressive} proposes a self-supervised correspondence learning method, which decomposes point clouds into sequences of semantically aligned shape compositions within a learned canonical space. Building upon this approach, we extract the key points of a point cloud in a canonical space with uniform semantic distribution.

\section{The Proposed Approach}
\label{approach}
\subsection{Preliminary}
We aim to address prior shape bias in masked autocoding and generative models for point clouds. Existing approaches have commonly resorted to leveraging farthest point sampling as a means to pre-sample center points. However, this method inadvertently confers the model with prior knowledge regarding the approximate shape of objects, thereby diminishing the inherent challenge of training the model. In contrast, our proposed approach introduces a differentiable center point acquisition method that fully leverages the semantic information of each part of the point cloud. This method effectively resolves the prior shape bias problem and significantly enhances the overall performance of the model. Figure \ref{framework} illustrates the architecture of our proposed method.
\subsection{Canonical Sphere Mapping}
\label{stage1}
In order to replace the traditional FPS for center point sampling and ensure that the proxy task for model training remains non-trivial, we aim to leverage the semantic information of different parts within the point cloud to compute the center points. However, the positions of different parts may vary for instances of one category, this can lead to semantic biases and uneven distributions. 

Motivated by \cite{cheng2021learning, cheng2022autoregressive}, we initialize a mapping function  $M(\cdot)$ with two MLPs that projects the point cloud onto a canonical sphere $\pi$. The canonical sphere can better capture the overall characteristics and structure of the point cloud data. By performing the reconstruction proxy task, corresponding parts of different instances in one category are mapped to the same region, effectively avoiding semantic biases. This ensures that the sphere representation reflects the inherent semantics of the point cloud data.

\begin{figure}
	\centering 
	\includegraphics[scale=0.41]{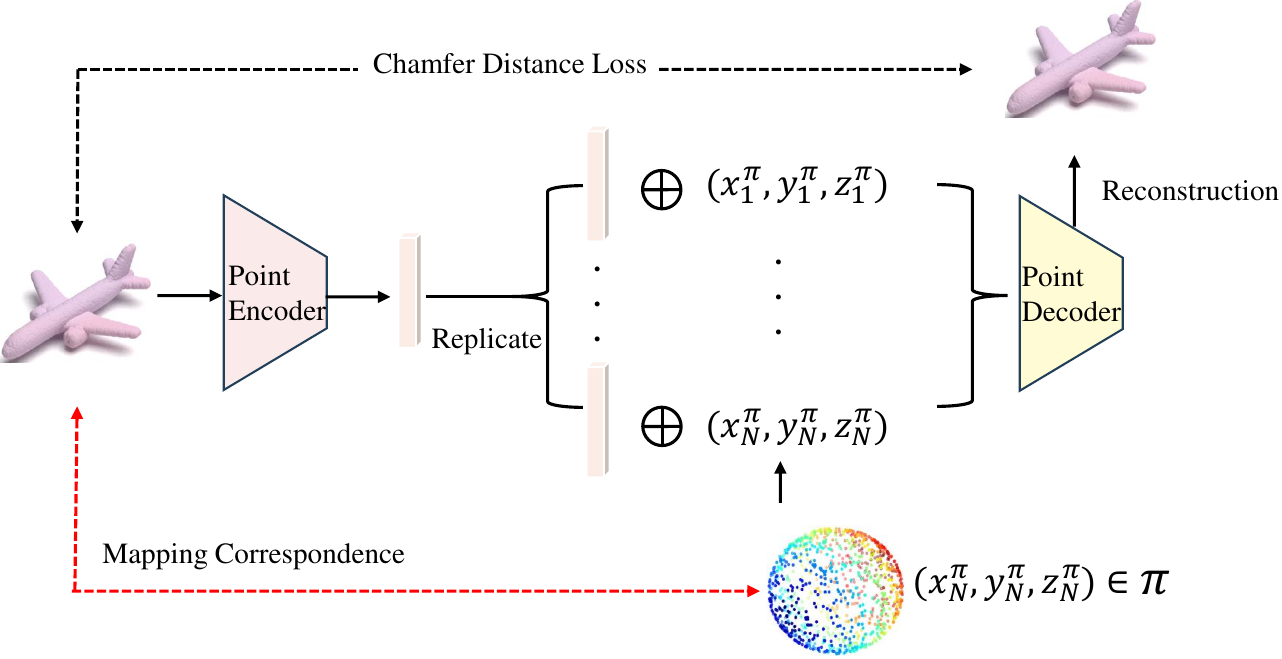} 
	\caption{\textbf{Overview of canonical mapping.} It consists of two key components: a point encoder responsible for generating shape features from the input point cloud, and a nonlinear decoder function that reconstructs the input shape using the canonical sphere and the extracted shape features.}  
	\label{cmap} 
\end{figure}

As depicted in Figure \ref{cmap}, given an input point cloud $p \in \mathbb R^{N \times 3}$ containing $N$ points, we employ DGCNN \cite{wang2019dynamic} to encode it into a latent embedding $l$. Then, we replicate multiple instances of the latent embedding and connect them with points $\{x_i^{\pi},y_i^{\pi},z_i^{\pi}\}$ sampled from the canonical sphere $\pi$, as the input of the nonlinear function $F(\cdot)$, initialized with two MLPs. At the output end of the function, we reconstruct the input point cloud using the Chamfer Distance loss \cite{fan2017point}:
\begin{equation}\label{1.5}
	\mathcal L_{\text{cd}}^1(\mathcal P,\mathcal G)\!=\!\frac{1}{\left| \mathcal P\right|}\sum_{p\in\mathcal P}\min\limits_{g \in \mathcal G}\Vert p-g \Vert+\frac{1}{\left| \mathcal G\right|}\sum_{g\in\mathcal G}\min\limits_{p \in \mathcal P}\Vert g-p \Vert,
\end{equation}
where $\mathcal P$ is the ground truth point set and $\mathcal G$ is the generated point set. In this way, we map the point cloud onto a canonical sphere with a more uniformly distributed semantic representation.

\subsection{Point Cloud Composition Learning}
\label{stage2}

Current approaches initially perform center point sampling using FPS in the point cloud. Each point is then assigned to the nearest center point, resulting in the partitioning of multiple patches. However, this approach leaks center point information during the generation process and does not consider semantic prior information,
often leading to discontinuous shape combinations.

To make the aforementioned process differentiable and consider the semantic information of the point cloud, we propose a differentiable approach to extract key points.  The semantic information on the canonical sphere is more uniformly distributed, and the positions of points on the sphere relative to the origin, as well as the distances between them, can provide additional  contextual information. Therefore, we compute a probability map $\mathcal Q \in \mathbb R^{N \times G}$ on the sphere, where $G$ denotes the number of groups and is set as 64 for all experiments conducted. Each group is referred to as a shape composition, which can be seen as an equivalent of an image patch in the 2D domain.

Inspired by \cite{cheng2022autoregressive}, we train a self-supervised composition learning network $U(\cdot)$. For each point on the sphere, we predict its composition assignment using MLP and a Softmax activation function. This results in the probability map $\mathcal Q$ of assignments for all points $\hat{p}$ in the canonical sphere, where $\mathcal Q_{i, j}$ represents the probability of assigning point $\hat{p_i}$ to the $j$-th group.

To ensure that the learned grouping can effectively formulate the overall semantic information of the whole point cloud, we compute $h$ composition points $\mathcal C \in \mathbb R^{h \times 3}$ analogous to center points, where each $\mathcal C_j$ is computed as follows:
\begin{equation}
	\mathcal C_j = \sum_{i=1}^{N}M^{-1}_{\pi \rightarrow p}(\hat{p_i})\cdot \mathcal Q_{i,j}, j \in \{1, 2, ..., h\},
\end{equation}
we let $M^{-1}_{\pi \rightarrow p}$ denote the inverse function of $M(\cdot)$, which maps points from the canonical sphere back to the original point cloud. Then we compute the Chamfer Distance loss $\mathcal L_{\text{CD}}^2$ between the predicted center points $\mathcal C$ and the ground truth point cloud $\mathcal P$:
\begin{equation}\label{1.5}
	\mathcal L_{\text{cd}}^2(\mathcal C,\mathcal P)\!=\!\frac{1}{h}\sum_{c\in\mathcal C}\min\limits_{p \in \mathcal P}\Vert c-p \Vert+\frac{1}{\left| \mathcal P\right|}\sum_{p\in\mathcal P}\min\limits_{c \in \mathcal C}\Vert p-c \Vert.
\end{equation}

This pretraining stage makes DCS-Net undergo a warm-up process and can be directly applied in more complex point cloud spaces without operating in the canonical sphere space.
\subsection{Differentiable Center Sampling Network (DCS-Net)}
The reconstruction of both the global and local patterns of the point cloud should be non-trivial. To simultaneously incorporate global feature reconstruction and local feature reconstruction as proxy tasks, we employ the pretrained composition learning network $U(\cdot)$ in section \ref{stage2} as a differentiable center sampling network (DCS-Net). Moreover, we employ the Gumbel softmax relaxation \cite{jang2016categorical} and a uniform prior for the probability map $\hat{\mathcal Q}$, allowing end-to-end training through backpropagation:
\begin{equation}\label{1.5}
	\mathcal Q = \text{DCS-Net}(p), \hat{\mathcal Q} = \text{Gumbel-softmax}(\mathcal Q).
\end{equation}
Considering the crucial semantic information in each part of the point cloud, we treat $\hat{\mathcal Q}$ as a weight map to perform weighted operations on the points in order to obtain the center points $\hat{\mathcal C} \in \mathbb R^{G \times 3}$:
\begin{equation}
	\hat{\mathcal C_j} = \sum_{i=1}^{N}p_i \cdot \hat{\mathcal Q}_{i,j}, j \in \{1, 2, ..., G\}.
\end{equation}

In order to realize the simultaneous learning of global pattern and local pattern in point cloud, we use global feature reconstruction and local feature reconstruction as proxy tasks. The local feature reconstruction task is handled by a subsequent point cloud model (such as the masked prediction task in GPM \cite{li2023general}), while the global feature reconstruction task is the center reconstruction task:
\begin{equation}\label{1.5}
	\mathcal L_{\text{cd}}^3(\hat{\mathcal C},\mathcal P)\!=\!\frac{1}{h}\sum_{c\in\hat{\mathcal C}}\min\limits_{p \in \mathcal P}\Vert c-p \Vert+\frac{1}{\left| \mathcal P\right|}\sum_{p\in\mathcal P}\min\limits_{c \in \hat{\mathcal C}}\Vert p-c \Vert.
\end{equation}
By employing DCS-Net as a replacement for FPS to obtain center points, we successfully address prior shape bias problem and enable both global and local patterns learning in the point cloud with non-trivial proxy tasks. Furthermore, it is plug-and-play for all point cloud models to extract centers and devise novel proxy tasks, such as mask center prediction.
\begin{table*}[h]\tiny
	\setlength{\belowdisplayskip}{-1.5cm}
	\centering
	\resizebox{\linewidth}{!}{
		\begin{tabular}{ccccc}
			\toprule
			\multirow{2}{*}{Methods} &\multicolumn{3}{c}{ScanObjectNN} & {ModelNet40}\\
			\cline{2-5} 
			& {OBJ\_BG} & {OBJ\_ONLY} & {PB\_T50\_RS}&{1k\_P}  \\
			\midrule
			\textit{\textbf{Self-Supervised Representation Learning}}\\
			PCT \cite{zhao2021point}&-&-&-&93.2 \\
			PointTransformer \cite{guo2021pct}&-&-&-&93.7 \\
			NPCT \cite{guo2021pct}&-&-&-&91.0 \\
			Transformer\cite{vaswani2017attention}&-&-&-&91.4 \\
			Transformer-OcCo\cite{vaswani2017attention}&-&-&-&92.1 \\
			\midrule
			Point-BERT\cite{yu2022point} & 87.4&88.1&83.1&92.8\\
			\rowcolor{gray!30}\textbf{Point-BERT+DCS-Net} &{\textbf{88.1}}
			&{\textbf{88.8}}&{\textbf{83.2}}&{\textbf{93.4}}\\
			\midrule
			MaskPoint\cite{liu2022masked} &88.4& 88.0&83.8 &92.9\\			
			\rowcolor{gray!30}\textbf{MaskPoint+DCS-Net} &{\textbf{88.6}}&{\textbf{88.4}}&{\textbf{84.1}} &{\textbf{93.1}}\\
			\midrule
			Point-MAE\cite{pang2022masked} &88.9&87.6&85.1&92.7\\
			\rowcolor{gray!30}\textbf{Point-MAE+DCS-Net} &{\textbf{89.2}}&{\textbf{88.0}}&\textbf{85.4}&{\textbf{92.8}}\\
			\midrule
			PointGPT\cite{chen2023pointgpt} &91.6&90.0&86.9&94.0\\
			\rowcolor{gray!30}\textbf{PointGPT+DCS-Net} &{\textbf{92.3}}&{\textbf{90.8}}&\textbf{87.1}&{\textbf{94.3}}\\
			\midrule
			PointM2AE\cite{zhang2022point}&91.22&88.81&86.43&92.4\\
			\rowcolor{gray!30}\textbf{PointM2AE+DCS-Net}& {\textbf{92.07}}&{\textbf{88.91}}&{\textbf{87.20}}&{\textbf{92.9}}\\
			MaskFeat3D\cite{yan20233d}&91.4&90.0&87.4&90.7\\
			\rowcolor{gray!30}\textbf{MaskFeat3D+DCS-Net}& {\textbf{92.4}}&{\textbf{90.8}}&{\textbf{88.6}}&{\textbf{91.6}}\\
			GPM \cite{li2023general}& 90.2&90.0&84.8&93.8\\
			\rowcolor{gray!30}\textbf{GPM+DCS-Net}& {\textbf{90.5}}&{\textbf{90.2}}&{\textbf{85.1}}&{\textbf{94.0}}\\
			\midrule
			\textit{\textbf{Methods with Cross-modal Information and Teacher Models}}\\
			ACT\cite{dong2022autoencoders} &{\textbf{93.3}}&91.9&88.2&93.7\\
			\rowcolor{gray!30}\textbf{ACT+DCS-Net} &93.1&{\textbf{92.3}}&{\textbf{88.5}}&{\textbf{94.0}}\\
			\midrule
			Recon\cite{qi2023contrast} &95.4&93.6&91.3&94.7\\
			\rowcolor{gray!30}\textbf{Recon+DCS-Net} &{\textbf{95.6}}&{\textbf{93.9}}&{\textbf{91.7}}&94.7\\
			\bottomrule
	\end{tabular}}
	\caption{\textbf{Classification results on the ScanObjectNN and ModelNet40 datasets.} The accuracy obtained on the ModelNet40 dataset is reported for 1k points. Overall accuracy (\%) are reported.}
	\label{1}
\end{table*}
\section{Experiment}
\label{experiment}
To validate the effectiveness of DCS-Net, we conduct experiments on existing point cloud models and multimodal point cloud models, including object classification, part segmentation, few-shot learning, and transfer learning (to validate the model's representation learning capabilities). Specifically, we replace the FPS algorithm with DCS-Net for differentiable center sampling, mitigating the issue of prior shape bias while enabling the model to learn the global pattern of the point cloud.
improves the performance of the model.
\begin{table}[h]\huge
	\centering
	\resizebox{9.0cm}{!}{
		\begin{tabular}{ccccc}
			\toprule
			\multirow{3}*{Methods} & \multicolumn{2}{c}{5-way} & \multicolumn{2}{c}{10-way}\\
			\cline{2-5}\\
			& 10-shot&20-shot&10-shot&20-shot\\
			\midrule
			\textit{\textbf{Self-Supervised Representation Learning}}\\
			Point-BERT \cite{yu2022point}&94.6 $\pm$ 3.1&96.3  $\pm$ 2.7&91.0  $\pm$ 5.4 & 92.7  $\pm$ 5.1\\
			\rowcolor{gray!30}\textbf{Point-BERT+DCS-Net}&{\textbf{96.4 $\pm$ 2.7}}&{\textbf{98.2$\pm$ 1.9}}&{\textbf{93.3$\pm$ 5.1}} & {\textbf{95.1$\pm$ 5.7}}\\	
			\midrule
			MaskPoint \cite{liu2022masked}&94.6  $\pm$ 4.1&96.8  $\pm$ 2.3&91.1  $\pm$4.2 & 92.9  $\pm$ 2.9\\
			\rowcolor{gray!30}\textbf{MaskPoint+DCS-Net}&{\textbf{96.6$\pm$ 2.8}}&{\textbf{98.4$\pm$ 3.2}}&{\textbf{92.8$\pm$ 4.9}}& {\textbf{94.6$\pm$ 4.1}}\\
			\midrule	
			Point-MAE \cite{pang2022masked}&94.9  $\pm$ 3.4&96.8 $\pm$ 2.2&91.8  $\pm$ 3.6 & 94.3  $\pm$ 2.7\\
			\rowcolor{gray!30}\textbf{Point-MAE+DCS-Net}&{\textbf{95.5$\pm$ 3.6}}&{\textbf{98.0$\pm$ 2.2}}&{\textbf{92.8$\pm$ 4.2}} &{\textbf{95.1$\pm$ 4.5}}\\
			\midrule
			PointGPT \cite{chen2023pointgpt}&96.8 $\pm$ 2.0&{\textbf{98.6$\pm$ 1.1}}&92.6  $\pm$ 4.6 & 95.2  $\pm$ 4.3\\
			\rowcolor{gray!30}\textbf{PointGPT+DCS-Net}&{\textbf{97.1$\pm$ 2.3}}&98.4  $\pm$ 0.9&{\textbf{92.8$\pm$ 4.3}}& {\textbf{95.4$\pm$ 4.1}}\\	
			PointM2AE \cite{zhang2022point}&96.8 $\pm$ 1.8&98.3$\pm$ 1.4&92.3  $\pm$ 4.5 & 95.0  $\pm$ 3.0\\
			\rowcolor{gray!30}\textbf{PointM2AE+DCS-Net}&{\textbf{97.2$\pm$ 2.0}}&{\textbf{98.5$\pm$ 3.7}}&{\textbf{92.8$\pm$ 4.0}}&{\textbf{95.3$\pm$ 3.1}}\\
			\midrule
			MaskFeat3D \cite{yan20233d}&97.1 $\pm$ 2.1&98.4$\pm$ 1.6&93.4  $\pm$ 3.8 & 95.7  $\pm$ 3.4\\
			\rowcolor{gray!30}\textbf{MaskFeat3D+DCS-Net}&{\textbf{97.6$\pm$ 1.9}}&98.4  $\pm$ 0.9&{\textbf{93.8$\pm$ 3.4}}& {\textbf{96.1$\pm$ 3.0}}\\	
			\midrule	
			GPM \cite{li2023general} & 97.0 $\pm$ 4.2&97.9 $\pm$ 3.2&92.6 $\pm$ 4.6&94.3 $\pm$ 5.4\\
			\rowcolor{gray!30}\textbf{GPM+DCS-Net}&{\textbf{97.3$\pm$ 4.0}}&{\textbf{98.9$\pm$ 2.8}}&{\textbf{93.0$\pm$ 4.7}} & {\textbf{95.1$\pm$ 4.7}}\\
			\midrule
			\textit{\textbf{Methods with Cross-modal Information and Teacher Models}}\\
			ACT\cite{dong2022autoencoders} &96.8 $\pm$ 2.3&98.0  $\pm$ 1.4&93.3  $\pm$ 4.0 & 95.6  $\pm$2.8\\
			\rowcolor{gray!30}\textbf{ACT+DCS-Net} &{\textbf{97.0$\pm$ 2.9}}&{\textbf{98.3$\pm$ 1.8}}&{\textbf{93.6$\pm$ 3.7}}& 95.6  $\pm$2.4\\
			\midrule
			Recon\cite{qi2023contrast} &97.3 $\pm$ 1.9&98.9  $\pm$ 1.2&93.8  $\pm$4.0 & {\textbf{95.8$\pm$ 2.8}}\\
			\rowcolor{gray!30}\textbf{Recon+DCS-Net} &{\textbf{97.7$\pm$ 1.8}}&{\textbf{99.0$\pm$ 1.0}}&{\textbf{94.0$\pm$ 3.3}}& 95.7  $\pm$ 2.7\\
			\bottomrule	\end{tabular}}
	\caption{\textbf{The results of few-shot classification on the ModelNet40 dataset.} For each experimental setting, we conduct 10 independent experiments and report the average accuracy (\%) along with its standard deviation.}
	\label{2}
\end{table}
\begin{table}[t]
		\centering
		\resizebox{\linewidth}{!}{
			\begin{tabular}{c|cc}
				\toprule
				{Methods} & Cls.mIoU & Inst.mIoU \\
				\midrule
				\textit{\textbf{Self-Supervised Representation Learning}}\\
				Point-BERT \cite{yu2022point} &84.11 &85.60\\
				\rowcolor{gray!30}\textbf{Point-BERT+DCS-Net}&{\textbf{84.24}} & 85.60\\
				\midrule
				MaskPoint \cite{liu2022masked}&83.4&85.1\\
				\rowcolor{gray!30}\textbf{MaskPoint+DCS-Net}&{\textbf{84.0}}&{\textbf{85.8}}\\
				\midrule
				Point-MAE \cite{pang2022masked}&83.6&85.66\\
				\rowcolor{gray!30}\textbf{Point-MAE+DCS-Net}&{\textbf{84.0}} & {\textbf{85.73}}\\
				\midrule
				PointM2AE \cite{zhang2022point}&84.9&86.5\\
				\rowcolor{gray!30}\textbf{PointM2AE+DCS-Net}&{\textbf{85.3}} & {\textbf{86.8}}\\
				\midrule
				MaskFeat3D \cite{yan20233d}&86.3&84.9\\
				\rowcolor{gray!30}\textbf{MaskFeat3D+DCS-Net}&{\textbf{86.6}} &{\textbf{85.4}}\\
				\midrule
				PointGPT \cite{chen2023pointgpt}&84.1&86.2\\
				\rowcolor{gray!30}\textbf{PointGPT+DCS-Net}&{\textbf{84.5}}& {\textbf{86.6}}\\
				\midrule
				GPM \cite{li2023general}&84.20&85.80\\
				\rowcolor{gray!30}\textbf{GPM+DCS-Net}&{\textbf{84.31}} &{\textbf{86.03}}\\
				\midrule
				\textit{\textbf{Methods with Cross-modal Information and Teacher Models}}\\
				ACT\cite{dong2022autoencoders} &84.7&86.1\\
				\rowcolor{gray!30}\textbf{ACT+DCS-Net} &{\textbf{85.0}}&{\textbf{86.5}}\\
				\midrule
				Recon\cite{qi2023contrast} &84.8&86.5\\
				\rowcolor{gray!30}\textbf{Recon+DCS-Net}&{\textbf{85.0}}&86.5\\
				\bottomrule
		\end{tabular}}
		\caption{\textbf{Part segmentation results on the ShapeNetPart dataset.} We report the average intersection mIoU over the union of all classes (Cls.) and instances (Inst.).}
		\label{3}
	\end{table}
	\subsection{Pretraining Dataset and Implementation}
	\label{pdl}
	Following the dataset configuration similar to existing models \cite{yu2022point, li2023general}, we employ ShapeNet \cite{chang2015shapenet} as our pretraining dataset, encompassing over 50,000 unique 3D models spanning 55 common object categories. We sample 1024 points from the point cloud and use DCS-Net instead of the FPS algorithm for center sampling. The point cloud is divided into $G=64$ groups, with each group containing 32 points. DCS-Net consists of two (convolutional layer, batch normalization layer) blocks with ReLU activation functions interspersed between them. We first pretrain DCS-Net following the guidelines in section \ref{stage2}. Then, we combine it with the point cloud model for further training, aiming to better learn the global patterns within the point cloud.
	\subsection{Downstream Tasks}
	\label{dt}
	\paragraph{Object Classification on Real-World Dataset.}
	We evaluate our \textbf{DCS-Net+pretrained model} on a challenging real-world dataset ScanObjectNN dataset \cite{uy2019revisiting}, which comprises approximately 15,000 objects extracted from real indoor scans, encompassing 2902 point clouds from 15 categories. The dataset poses a more significant challenge due to the inclusion of real-world scans with backgrounds and occlusions. Following prior work \cite{yu2022point, li2023general, chen2023pointgpt, pang2022masked}, we conduct experiments on three main variants: OBJ-BG, OBJ-ONLY, and PB-T50-RS. Different from the settings in Point-MAE \cite{pang2022masked} and PointGPT \cite{chen2023pointgpt}, we follow the configuration used in Point-BERT \cite{yu2022point} and uniformly partition the point cloud into 64 groups. 
	\begin{figure}
		\centering 
		\includegraphics[scale=0.4]{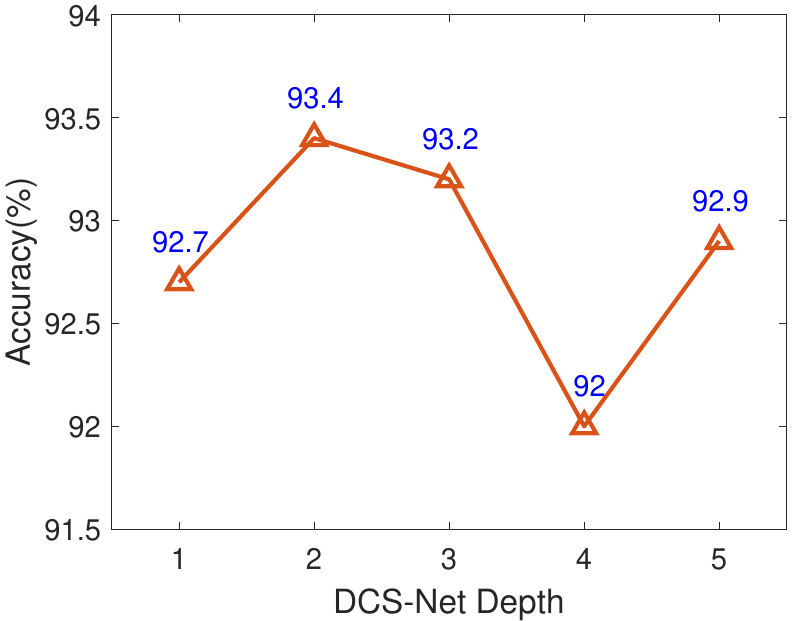} 
		\caption{\textbf{Ablation study of DCS-Net depth.} The DCS-Net depth represents the ablation for classification finetune on ModelNet40 dataset. We conduct experiments with depth = $\{1,2,3,4,5\}$, and when the depth is 2, the finetuning effect is the best.}  
		\label{ab1} 
	\end{figure}
	In addition to validating our approach on the classical self-supervised point cloud model, we conduct relevant experiments on two multimodal point cloud models. From the experimental results in Table \ref{1}, it can be observed that DCS-Net not only addresses the issue of prior shape bias in traditional models but also achieves moderate improvements in the evaluation metrics. This demonstrates the effectiveness of our approach in capturing more accurate representations of the global patterns in point clouds.
	\paragraph{Object Classification on Clean Object Dataset.}
	We evaluate our method on the ModelNet40 dataset \cite{wu20153d}, comprising 12,311 clean 3D CAD models spanning 40 categories. Following the experimental setup of Point-BERT \cite{yu2022point}, we utilize a two-layer MLP with a dropout rate of 0.5 as the classification head for the task. The model optimization is performed using AdamW with a weight decay of 0.05 and a learning rate of 0.0005. A batch size of 32 and a cosine annealing schedule are employed. Additionally, we only report the accuracy achieved on the ModelNet40 dataset for 1k points.The results presented in Table \ref{1} demonstrate that our method obtains more competitive results.
	\paragraph{Few-shot Learning.}
	To demonstrate the model's ability of acquiring knowledge for new tasks with limited training data, we evaluate it in the context of few-shot learning. Following the typical '$W$-way $S$-shot' setup, we randomly select $W$ classes and sample ($S$+20) objects for each class \cite{sharma2020self}. The model is then trained on $W \times S$ samples (support set) and evaluated on the remaining 20$W$ samples (query set). We conduct 10 independent experiments for each setting and report the average performance and standard deviation across the 10 runs. As shown in Table \ref{2}, DCS-Net enhances the model's generalization capability by employing both global feature reconstruction and local feature reconstruction as proxy tasks.
	
	\paragraph{Part Segmentation.}
	We assess the representation learning capability of our approach on the ShapeNetPart dataset \cite{yi2016scalable}, with the objective of predicting more fine-grained class labels for each point. This dataset comprises 16 categories and includes a total of 16,881 objects. We sample 2048 points from each model and increase the $G$ from 64 to 128. The segmentation head \cite{pang2022masked} connects features $\mathcal F^4, \mathcal F^8, \mathcal F^{12}$ extracted from the 4-$th$, 6-$th$, and 12-$th$ layers of the transformer blocks. Subsequently, average pooling, max pooling, and upsampling are employed to generate features for each point, followed by label prediction using a multi-layer perceptron (MLP). The experimental results presented in Table \ref{3} demonstrate the superior performance of our DCS-Net.
	\subsection{Real-world Object detection}
	To validate the efficacy of DCS-Net in the context of point cloud object detection, we conduct experiments on the ONCE \cite{mao2021one} dataset, shown in Table \ref{obj_zheng}. ONCE dataset incorporates three types of coordinate systems: the LiDAR coordinate, the camera coordinates, and the image coordinate. The LiDAR coordinate system is centered at the LiDAR sensor, with the x-axis pointing to the left, the y-axis pointing backwards, and the z-axis pointing upwards. The camera coordinates can be directly converted to the LiDAR coordinate system using the respective extrinsics. The image coordinate system is a 2D system where the origin is located at the top-left corner of the image. The x-axis runs along the width of the image, and the y-axis runs along the height. The transformation from the camera coordinate to the image coordinate is achieved using the camera intrinsics. 
	
	We conduct the expermient on the PointRCNN \cite{shi2019pointrcnn} and PV-RCNN \cite{shi2020pv}. The metrics used for evaluation include mean Average Precision (mAP) on vehicle, predestrian, cyclist and average of them.  Unlike other methods that require both RGB images and point clouds as inputs, PointRCNN and PV-RCNN only receive point clouds as input. Here we replace the FPS center point sampling method in the two models with DCS-Net and achieve better detection performance, thereby validating the effectiveness of DCS-Net in the real-world object detection dataset.


	\subsection{Real-world Segmentation}
	To validate the efficacy of DCS-Net in the context of real-world segmentation, we conduct experiments on the S3DIS \cite{armeni20163d} dataset in Point Transformer \cite{zhao2021point} and StratifiedFormer \cite{lai2022stratified}, the results are shown in Table \ref{seee_zheng}. S3DIS dataset is utilized for semantic scene parsing, encompasses 271 individual rooms distributed across six distinct areas within three separate buildings. Each point within the scan is assigned a semantic label from one of 13 predefined categories, such as ceiling, floor, and table. 
	
	We evaluate the Point Transformer  and StratifiedFormer by excluding Area 5 during the training phase and then testing on it. The metrics used for evaluation include mean class-wise intersection over union (mIoU), mean class-wise accuracy (mAcc), and overall point-wise accuracy (OA). Similar to the detection task, we replace the FPS sampling process in Point Transformer and StratifiedFormer with the differentiable center point sampling (DCS-Net), achieving moderate improvements and reducing the shape prior bias problem. The parameter settings during the experiment are consistent with those in the original paper. 

		\begin{table}[t]
		\centering
		\makeatletter\def\@captype{table}
		\resizebox{\linewidth}{!}{
			\begin{tabular}{ccccc}
				\toprule
				{Methods}&Vehicle (\%)&Pedestrian (\%)&Cyclist (\%)&Average (\%)\\
				\midrule
				PointRCNN \cite{shi2019pointrcnn}&52.09&4.28&29.84&28.74\\
				\rowcolor{gray!30}PointRCNN+DCS-Net&62.74&10.29&35.41&32.90\\
				\midrule
				PV-RCNN \cite{shi2020pv}&77.77&23.50&59.37&53.55\\
				\rowcolor{gray!30}PV-RCNN+DCS-Net&79.63&25.60&62.92&54.39\\
				\bottomrule
		\end{tabular}}
		\caption{\textbf{Real-world object detection results on ONCE dataset.} We report the mAP (\%) on vehicle, pedestrian, cyclist and average of them to validate the effectiveness of our approach.}
		\label{obj_zheng}
	\end{table}
	\begin{table}[t]
		\centering
		\makeatletter\def\@captype{table}
		\resizebox{\linewidth}{!}{
			\begin{tabular}{cccc}
				\toprule
				{Methods}&mIOU (\%)&mACC (\%)&OA (\%)\\
				\midrule
				Point Transformer \cite{zhao2021point}&70.4&76.5&90.8\\
				\rowcolor{gray!30}Point Transformer+DCS-Net&70.8&77.0&91.0\\
				\midrule
				StratifiedFormer \cite{lai2022stratified}&71.7&77.6&91.9\\
				\rowcolor{gray!30}StratifiedFormer+DCS-Net&72.1&77.9&92.1\\
				\bottomrule
		\end{tabular}}
		\caption{\textbf{Real-world segmentation results on S3DIS dataset.} We report the mIOU (\%), mACC (\%) and OA (\%) as the main metric. The results show that DCS-Net can not only address the shape prior bias, but improve the performence of model on the real-world segmentation.}
		\label{seee_zheng}
	\end{table}
	\subsection{Ablation Study}
	\paragraph{Hyper Parameter.}
	Figure \ref{ab1} demonstrates the ablation study on the DCS-Net depth during Point-BERT pretraining and whether to finetune the DCS-Net during the fine-tuning phase. It can be observed that the DCS-Net achieves optimal performance when the network depth is set to 2. Additionally, applying the stop-gradient operation to the DCS-Net during finetuning not only aligns with the plug-and-play requirement but also leads to improved performance.

	\paragraph{Reconstruction Loss.}
	By leveraging DCS-Net to acquire center points, we enable the model to treat center reconstruction as a proxy task and learn the global patterns of point clouds. Table \ref{7} presents the performance of variants using different center reconstruction loss functions, including $l$1-form CD loss, $l$2-form CD loss, Maximum Mean Discrepancy (MMD) loss, and a combination of both $l$1 and $l$2-forms CD loss, in terms of classification results on the ModelNet40 dataset. MMD is primarily used to measure the distance between two different yet related distributions:
	\begin{equation}\label{1.5}
		\mathcal L_{\text{MMD}}({S_g},S_r)=\frac{1}{|S_r|}\sum_{X \in S_r}\min\limits_{Y \in S_g}D(X,Y),
	\end{equation}
	where $S_g$ denotes the generated set and $S_r$ represents the query set. The function $D(X,Y)$ can either be Chamfer Distance or Earth Mover's Distance. Moreover, we conduct many additional ablation studies and discussions in the supplementary, including reconstruction loss in canonical mapping and composition learning, whether or not freeze DCS-Net during finetuning and so on. 
	\begin{table}[h]
			\centering
			\setlength{\tabcolsep}{1.5mm}{
				\begin{tabular}{cc>{\columncolor{gray!30}}ccc}
					\toprule
					{Model (\textit{with DCS-Net})} & $l$1 CD & $l$2 CD& $l$1+$l$2 CD&MMD\\
					\midrule
					Point-BERT&93.1&93.4&93.2&93.2\\
					MaskPoint&92.7&93.1&92.9&92.4\\
					Point-MAE&92.5&92.8&92.8&91.9\\
					Point-M2AE&92.6&92.9&92.8&92.0\\
					MaskFeat3D&91.4&91.6&91.4&90.8\\
					PointGPT&92.9&93.3&93.1&92.7\\
					GPM&93.4&94.0&93.6&93.4\\
					Recon&93.6&94.0&93.7&93.4\\
					ACT&94.1&94.7&94.4&94.1\\
					\bottomrule
			\end{tabular}}
			\caption{\textbf{Ablation study on the global feature reconstruction loss.} We report the accuracy achieved through finetuning on the ModelNet40 dataset. }
			\label{7}
		\end{table}
		
\section{Conclusion}
	In this study, we present \textbf{DCS-Net}, an innovative pretraining framework specifically designed for point cloud models, which effectively eliminates prior shape bias issue and facilitates the exploration of global patterns. DCS-Net introduces a differentiable approach to center point acquisition, which enhances the complexity of the training process. By incorporating non-trivial proxy tasks for global and local feature reconstruction, the model successfully captures both global and local patterns within the point cloud. The experimental results demonstrate that replacing the original farthest point sampling (FPS) algorithm with DCS-Net for center point acquisition not only mitigates prior shape bias in the point cloud model but also significantly improves performance in downstream point cloud tasks. This highlights the effectiveness and potential of DCS-Net in enhancing the capabilities of point cloud models for various applications.

\bibliography{aaai25}

\end{document}